# DIVERSITY AND INTELLIGENCE IN MULTI-ROBOT TEAMS

**Monica Dragoicea**

*University "Politehnica" Bucharest, Faculty of Control and Computers*
*Automatic Control and System Engineering Department*
*313 Splaiul Independentei, 060042 - Bucharest, Romania*
*E-mail: ma-dragoicea@ics.pub.ro*

Abstract: This research proposes new tools for investigation of behavioral diversity in multi-robot systems and a significant body of results using these tools in simulated and real mobile robot experiments. The experiments specifically describe a framework of defining behavior-based strategies for multi-robot tasks as robot foraging, robot soccer and robot formation. The research focuses specifically on motor schema-based multi-robot systems, which are an important example of behavior-based control..

## 1. INTRODUCTION

Multi-agent teams are desirable for many reasons. In the case of planetary explorers or removing land mines, more robots should be able to cover more area. Like ants and other insects, many cheap robots working together could replace a single expensive robot, making multi-agents more cost effective. Multi-agent teams are becoming quite popular in robot competitions, where teams of real or simulated robots play soccer against other teams. The soccer task explicitly requires multiple robots that must cooperate with each other, yet react as individuals.

In multi-agent systems the concurrent but independent actions of each robot leads to an *emergent social behavior*. The group behavior can be different from the individual behavior, emulating a certain *group dynamics*. Therefore designing teams is hard. Questions like "how does a designer recognize the characteristics of a problem that make it suitable for multi-agents?", "how does the designer (or the agents themselves) divide up the task?", "are there any tools to predict and verify the social behavior?" must be answered.

The whole field of multi-agents is so new that there is no consensus on what are the important dimensions and characteristics in describing a team. In general, *heterogeneity*, *control*, *cooperation* and *goals* could be used as these dimensions.

The three multi-robot tasks - *robotic foraging, robot soccer* and *formation maintenance* - benefit from a spectrum bounded by centralized control and distributed control regimes. The control strategy of the mobile robots must be based on reactive and goal-achieving procedures, which includes the perception, learning, planning and behavior generation phases. Therefore, the integration of advanced environment perception and communication devices into mechatronic structures facilitates the development of strongly associative information systems with high level of intelligence (Dumitrache, 2004).

## 2. HOMOGENOUS TEAMS OF ROBOTS

Heterogeneity refers to the degree of similarity between individual robots that work within a collection.

Most research in multi-robot teams to date has centered on homogenous systems, with work in heterogenous groups focused primarily on mechanical and sensor differences between agents. In contrast, this work examines teams of mechanically identical robots. These systems are interesting because they may be homogenous or heterogenous depending only on behavior. In homogenous teams the members are all identical. But members can be homogenous for one portion of a task by running identical behavior, then become heterogenous if the team members change the behavioral mix or tasks.

Robot foraging and robot soccer are two examples of robot tasks that describe artificial societies in which societal rules can be analyzed.

The forage task involves a collection of objects of interest (attractors) scattered about the environment. In a typical way, an agent begins by wandering about the environment looking for attractors. When an attractor is encountered, the robot moves towards it and grasp it. After attachment, the robot returns the object to a home base. Foraging has a strong biological basis and is an important subject of research in the mobile robotics community (Arkin, 1992), (Goldberg, 1997), (Fontan, 1997).

Robotic soccer is a particularly good task for multi-agent research because it includes cooperation between teammates, competition versus an opponent and unpredictable dynamic play. Recent interest has sparked more research in robot soccer. Kitano and Asada promote the Robot World Cup as a vehicle for multi-agent research (Kitano, 1997). Both robot foraging and robotic soccer *exhibit active cooperation* as well as *non-active cooperation* in pursuing their tasks.

## 3. FSM FOR TEAM DESIGN

The experiments presented in this work specifically describe a framework of defining behavior-based strategies for multi-robot tasks as robot foraging, robot soccer and robot formation. The research focuses specifically on motor schema-based multi-robot systems, which are an important example of behavior-based control. Individual motor schemas, or primitive behaviors, express separate goals or constraints for a task. Motor schemas may be grouped to form more complex, emergent behaviors. Groups of behaviors are referred to as *behavioral assemblages*.

One way behavioral assemblages may be used in solving complex tasks is to develop an assemblage for each sub-task and to execute the assemblage in an appropriate sequence, by *temporal sequencing* (Balch, 1998). Therefore a resulting task solving strategy can be represented as Finite State Machine. In order to solve a specific task, a robot may be in a certain state (e.g. wander, acquire, deliver, etc). In this work, for sequential real-time system we use a Finite State Machine (FSM) representation which represent the different states of the robot.

### 3.1 Homogenous foraging

Figure 1 displays an example of behavioral assemblage for a homogenous foraging task, that means all the robots collect all the different types of attractor and deliver them to the corresponding color-coded delivery areas. All agents are programmed with the same sequence of behaviors. Various other strategies (e.g. foraging specialized by color, territorial specialization) could be all built from the same repertoire of behaviors (Balch, 1998).

When activated (starting from OFF state), each robot starts to WANDER, roaming about the environment in search from attractors.

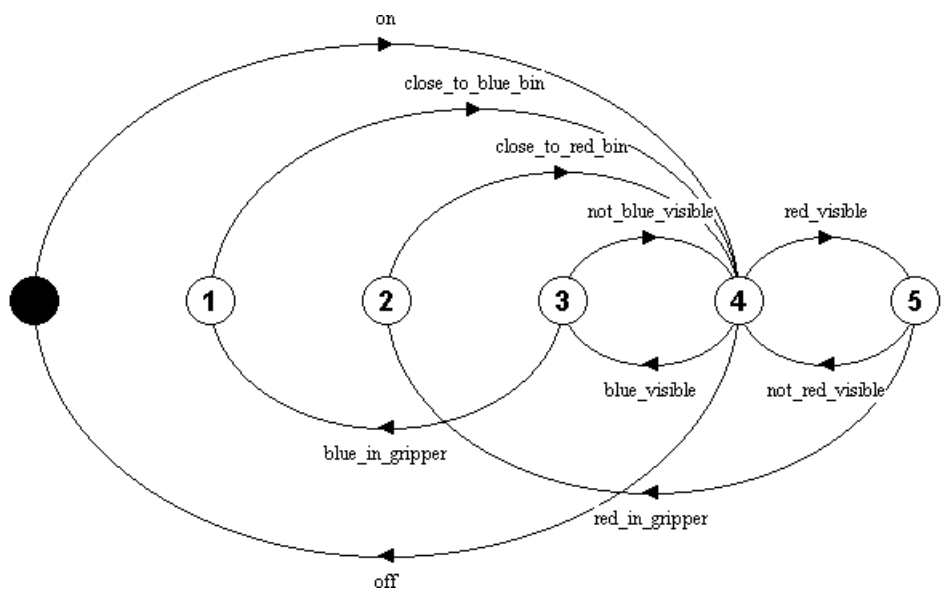

Fig. 1. Behavioral assemblage for homogenous foraging

As perceptual triggers for transitions to ACQUIRE_RED or ACQUIRE_BLUE behaviors special features are used, as red_visible and blue_visible releasers. Upon capturing an attractor, a robot returns back to home base using one of the DELIVER_RED or DELIVER_BLUE behaviors.

The behavior assemblage for homogenous foraging can be represented as an abstract model as follows:

```
HOM_FOR       = OFF,
OFF           = (on->WANDER),
WANDER        = (red_visible->ACQUIRE_RED
                |blue_visible->ACQUIRE_BLUE
                |off->OFF),
ACQUIRE_RED   = (red_in_gripper->DELIVER_RED
                |not_red_visible->WANDER),
DELIVER_RED   = (close_to_red_bin->WANDER),
ACQUIRE_BLUE  = (blue_in_gripper->DELIVER_BLUE
                |not_blue_visible->WANDER),
DELIVER_BLUE  = (close_to_blue_bin->WANDER).
```

Therefore, the Finite State Machine (FSM) that describes the homogenous foraging task is an extension of the behavioral table (see table 1):

$$M: (K, \Sigma, \delta, s, F)$$

where

$$K = \{off, WANDER, ACQUIRE\_RED, ACQUIRE\_BLUE, DELIVER\_RED, DELIVER\_BLUE\}$$

are the states the robot should be in (the finite number of discrete states for homogenous foraging is 6).

$$\Sigma = \{on, red\_visible, blue\_visible, off, red\_in\_gripper, not\_red\_visible, close\_to\_red\_bin, blue\_in\_gripper, not\_blue\_visible, close\_to\_blue\_bin\}$$

is the set of behavioral releasers (that means the inputs of the FSA, also called the alphabet).

$\delta$ is the transition function that specifies what state the robot is in after it encounters an input stimulus from $\Sigma$.

$$s = OFF$$

is the Start State, and the robot should always start there.

$$F = OFF$$

is the final state that the robot can reach that terminates the task. Here the final state is OFF, that means the robot runs the sequence of behaviors until it is turned off manually.

### 3.2 ROBOTIC SOCCER

Robotic soccer implies certain tasks in multi-agent systems that requires both active cooperation and non-active cooperation. In active cooperation one robot can pass the ball to another robot as part of an offensive play. The cooperation does not require communication - if a robot has the ball, can't see goal and can see team mate, then it passes to team mate, but this does require being aware of the teammates.

In non-active cooperation robots are programmed to individually pursue a goal without acknowledging other robots but cooperation emerges. Non-active cooperation has attracted much interest in the robotic community because it requires very little sensing or behaviors.

The choice of cooperation schemes is often influenced by the sensory capabilities of the robots and leads to different behavioral assemblages that fulfill the robot tasks. This work proposes a framework in which behaviors are defined, grouped and analyzed for feasibility in behavioral assemblages for robot soccer tasks.

Robot soccer is more than one robot generation beyond simpler competitions like solving a maze. In soccer, not only do we have a lack of environment structure (less walls), but we now have teams of robots playing an opposing team, involving moving targets (ball and other players), requiring planning, tactics, and strategy - all in real-time. So, obviously, this opens up a whole new dimension of problem categories. Robot soccer will remain a great challenge for years to come.

Behavior-based approaches are well suited for robot soccer since they excel in dynamic and uncertain environments. The robot assemblages of behaviors for robot soccer are described as Finite State Machines (see figure 2) and implemented using the potential field method.

In order to assure consistent results, the members of the team will always follow a fixed policy in terms of sequence of behaviors. Each robot selects from a set of behavioral assemblages to complete the task. The behaviors are sequenced to form a complete strategy.

The behavioral assemblages developed in this work, and the motor schemas activated are the following (see table 1):

- *goal-keeper*, has two elementary behaviors, triggered based on the ball position on the field
    - DEFEND: the robot always tries to get to the half distance between ball and the middle of its team gate. This behavior is activated when the ball is relatively far from the gate. The robot feels a repulsive force generated by the other robots and by the walls

- GO_TO_BALL: elementary behavior of the robots which is activated when the ball is close to the gate. The robot is attracted by the ball and feels a repulsive force generated by the other robots and the walls.
- *forward*, has two elementary behaviors, triggered based on their position on the field and the position of the ball:
  - BEHIND_BALL: this behavior is activated when the robot is close to the ball, being positioned between the ball and the opponent gate
  - GO_TO_BALL: this is the same behavior the goal-keeper has, the robot being attracted by the ball.

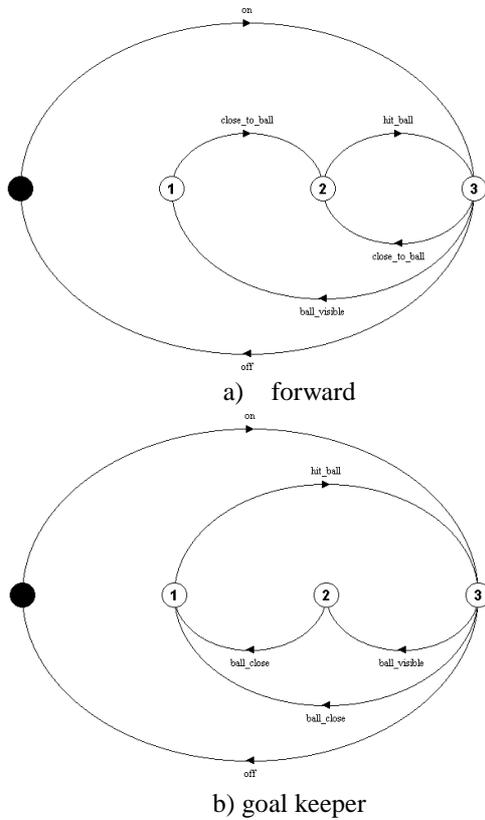

a) forward

b) goal keeper

Fig. 2. Behavioral assemblage for robotic soccer.

All the other field players (center half, outside left, outside right) have the same behaviors, being different only on their position on the playground (see figure 3). For example, in one team, the *forward* player activates only on (1,1), (1,2) and (1,3) sectors, the *outside left* player on (2,3) and (3,3), the *outside right* player on (2,1) and (3,1), the *center half* player on ((2,2), and the *goal keeper* on (3,2) sector.

### 4. MOTOR SCHEMAS FORMULATION AND IMPLEMENTATION

Behaviors composing a behavioral assemblage represented by a FSM, as well as releasers, can be designed using Object Oriented Principles - OOP, based on the motor schema approach proposed by (Arbib, 1991). Motor schemas can be implemented using a potential field approach.

The potential field method is widely used for autonomous mobile robot path planning due its elegant mathematical analysis and simplicity (Latombe, 1991). However, many researches focused on solving the motion planning problem in stationary environment where both targets and obstacles are stationary. For the implementation of the motor schema in the proposed mobile robot soccer strategies this work use a new potential field method for motion planning of the mobile robots in a dynamic environment where the target and the obstacles are moving (Ge, 2002).

These new potential functions take into account not only the relative positions of the robot with respect to the target and obstacles, but also the relative velocities of the robot with respect to the target and obstacles.

Therefore, when the target is moving, the conventional pure position based potential function is not directly applicable and has to be modified. The new attractive potential field function can be selected as follows (Ge, 2002):

$$U_{att}(\underline{p},\underline{v}) = \alpha_p \left\| \underline{p}_{tar}(t) - \underline{p}(t) \right\|^m + \alpha_v \left\| \underline{v}_{tar}(t) - \underline{v}(t) \right\|^n$$
(1)

where $\underline{p}(t)$ and $\underline{p}_{tar}(t)$ are the positions of the robot and the target at time t, respectively; $\underline{v}(t)$ and $\underline{v}_{tar}(t)$ are the velocities of the robot and target at time t, respectively; $\alpha_p$ and $\alpha_v$ are scalar positive parameters for the Euclidean distance between the robot and the target at time t, and for the magnitude of the relative velocity between the target and the robot at time t, respectively. M and n are positive constants.

Accordingly, the virtual force acting and driving the robot is defined as the negative gradient of the potential with respect to both position and velocity and is defined as follows:

$$F_{att}(\underline{p}) = -\nabla U_{att}(\underline{p}) = -\frac{\partial U_{att}(\underline{p})}{\partial \underline{p}}$$
(2)

Figure 4 shows the total virtual force acting on the robot while it tries to reach the ball in order to "save" its own gate. The red team robot has a BEHIND_BALL behavior because its position is close to the ball, between ball and the opponent gate.

It has to go behind ball in order to reach a favorable position to hit the ball to the right side of the field. This robot will obtain a trajectory that makes it to increase the distance to the ball.

Table 1 Behavioral assemblages for robot tasks

| Process | Homogenous Foraging | Robot Soccer | |
| --- | --- | --- | --- |
| | | Forward | Goal-keeper |
| States # | 6 | 4 | 4 |
| Alphabet | on<br>red_visible<br>blue_visible<br>off<br>red_in_gripper<br>not_red_visible<br>close_to_red_bin<br>blue_in_gripper<br>not_blue_visible<br>close_to_blue_bin | ball_visible, close_to_ball<br>hit_ball<br>off<br>on | on<br>ball_visible<br>ball_close<br>off<br>hit_ball |
| Transitions | 0 on 4<br>1 close_to_blue_bin 4<br>2 close_to_red_bin 4<br>3 blue_in_gripper 1<br>3 not_blue_visible 4<br>4 red_visible 5<br>4 blue_visible 3<br>4 off 0<br>5 red_in_gripper 2<br>5 not_red_visible 4 | 0 on 3<br>1 close_to_ball 2<br>2 hit_ball 3<br>3 ball_visible 1<br>3 close_to_ball 2<br>3 off 0 | 0 on 3<br>1 hit_ball 3<br>2 ball_close 1<br>3 ball_visible 2<br>3 ball_close 1<br>3 off 0 |
| States | 0 = OFF<br>1 = DELIVER_BLUE<br>2 = DELIVER_RED<br>3 = ACQUIRE_BLUE<br>4 = WANDER<br>5 = ACQUIRE_RED | 0 = OFF<br>1 = GO_TO_BALL<br>2 = BEHIND_BALL<br>3 = WANDER | 0 = OFF<br>1 = GO_TO_BALL<br>2 = DEFEND<br>3 = WANDER |

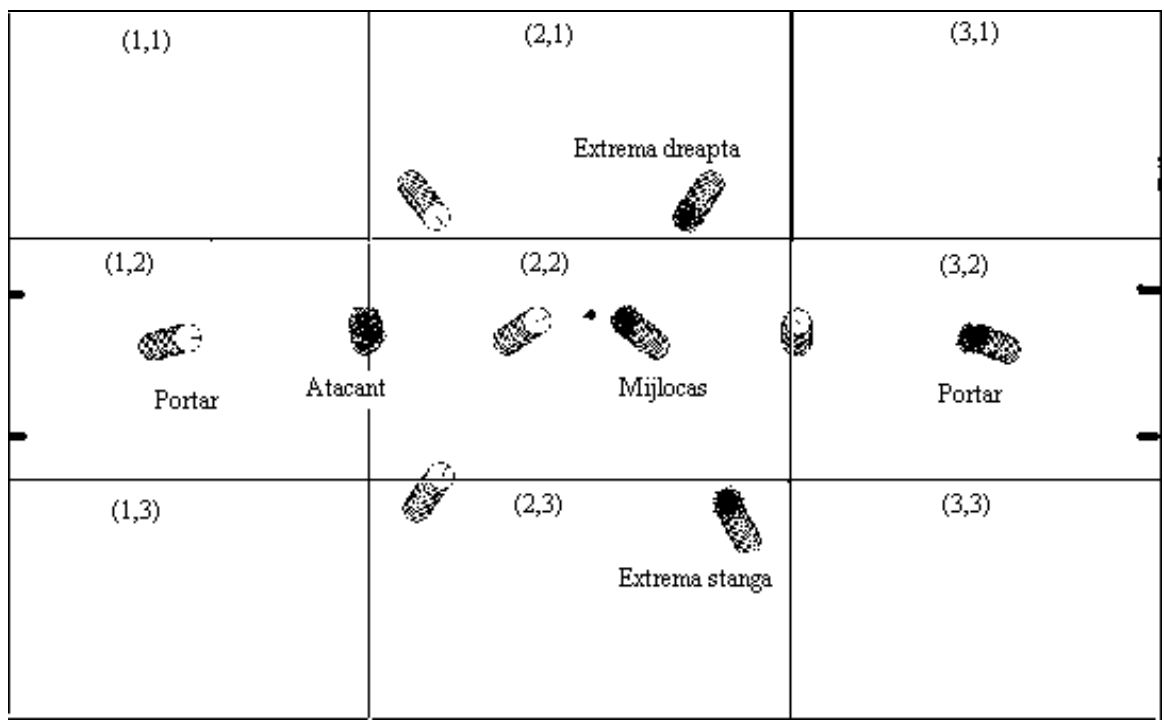

Fig. 3. Playground for robot soccer

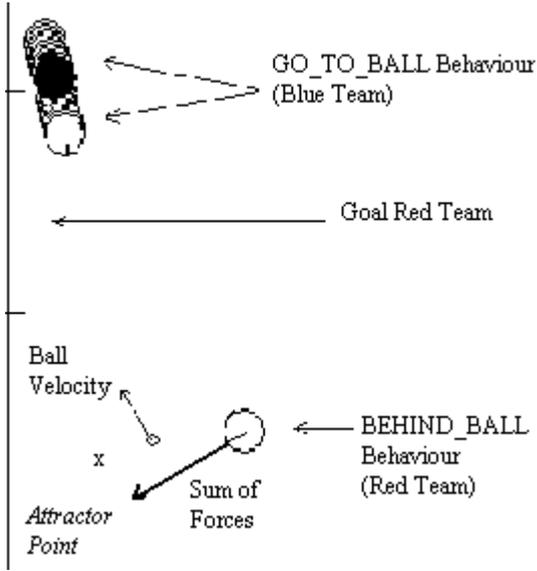

Fig. 4. Total virtual force for attack

The repulsive potential field function makes use of the position and velocity information of the obstacle. The relative velocity between the robot and the obstacle in the direction from the robot to the obstacle is given by:

$$v_{RO}(t) = [\underline{v}(t) - \underline{v}_{obs}(t)]^T \underline{n}_{RO} \quad (3)$$

where $\underline{n}_{RO}$ is a unit vector pointing from the robot to the obstacle. Accordingly, the repulsive potential can be defined as follows:

$$U_{rep}(\underline{p},\underline{v}) = \quad (4)$$

$$= \begin{cases} 0, & \text{if } \rho_s(p,p_{obs}) - \rho_m(v_{RO}) \geq \rho_0 \text{ or } v_{RO} \leq 0 \\ \eta\left(\dfrac{1}{\rho_s(p,p_{obs}) - \rho_m(v_{RO})} - \dfrac{1}{\rho_0}\right) & \text{if } 0 < \rho_s(p,p_{obs}) - \rho_m(v_{RO}) < \rho_0 \\ & \text{and } v_{RO} > 0 \\ \text{not defined}, & \text{if } v_{RO} > 0 \text{ and } \rho_s(p,p_{obs}) < \rho_m(v_{RO}) \end{cases}$$

Equations (1) to (4) can be used in order to implement the motor schemes of the elementary behaviors, by combing attractive and repulsive potential fields.

## 5. CONCLUSIONS

Multi-robot team design is challenging because performance depends significantly on issues that arise solely from interaction between agents. These interactions complicate development since they aren't obvious in the hardware or software design but only emerge in an operating team. *Cooperation, robot-robot interference* and *communication*, for instance, are not considerations for a single robot, but are crucial in multi-robot systems. Still automatic methods for matching multi-robot configuration to task don't yet exist; in most cases multi-agent design is ad-hoc.

This research seeks to address that by applying a principled approach to the analysis and design of behavior-based multi-robot teams.